\documentclass[12pt]{article}

\usepackage{natbib,hyperref}

\usepackage{amsmath}
\usepackage{amssymb}
\usepackage{graphicx}
\usepackage{float}

\usepackage{times}

\newenvironment{sciabstract}{%
\begin{quote} \bf}
{\end{quote}}

\newcounter{lastnote}

\title{Symphony from Synapses: Neocortex as a Universal Dynamical Systems Modeller using Hierarchical Temporal Memory}

\author
{Fergal Byrne \\
\normalsize{HTM Theory Group, Dublin, Ireland}\\
\\
\normalsize{fergal@brenter.ie http://inbits.com}
}

\date{}

\begin{document} 


\baselineskip14pt


\maketitle


\begin{sciabstract}
  Reverse engineering the brain is proving difficult, perhaps impossible. While many believe that this is just a matter of
  time and effort, a different approach might help. Here, we describe a very simple idea which explains the 
  power of the brain as well as its structure, exploiting complex dynamics rather than abstracting it away. 
  
  Just as a Turing Machine is a Universal 
  Digital Computer operating in a world of symbols, we propose that the brain is a Universal Dynamical Systems Modeller, evolved bottom-up 
  (itself using nested networks of interconnected, self-organised dynamical systems) to prosper in a world of dynamical systems.
  
  Recent progress in Applied Mathematics has produced startling evidence of what happens when abstract Dynamical Systems 
  interact. Key latent information describing system A can be extracted by system B from very simple signals, and signals can be used 
  by one system to control and manipulate others.
  Using these facts, we show how a region of the neocortex uses its dynamics to intrinsically ``compute'' about the external
  and internal world.
  
  Building on an existing ``static'' model of cortical computation (Hawkins' Hierarchical Temporal Memory - HTM), we describe how a region of neocortex can be viewed as a
  network of components which together form a Dynamical Systems modelling module, connected via sensory and motor pathways to
  the external world, and forming part of a larger dynamical network in the brain. 
    
  Empirical modelling and simulations of Dynamical HTM are possible with simple extensions and combinations of currently existing open source 
  software. We list a number of relevant projects. 
   
\end{sciabstract}

\newpage
\section*{Introduction}
  The mammalian brain is a complex network of interconnected dynamical systems, a property repeated at every scale, down to 
  the individual synapse and beyond. While this is also true of digital computers in the physics of components and circuits, we
  use engineering to design this away, building deterministic, precisely known mechanisms in successive ``logical'' layers, 
  in hardware and software. 
  Given the readily accessible concepts and tools provided by Computer Science, Applied Mathematics, Symbolic Logic, Information Theory and 
  Statistics, it is tempting to treat the brain as an information-processing system of obscure design, whose details might be revealed 
  by ``reverse-engineering'' (inferring a mechanism from knowledge of its functional behaviour). This is indeed the approach taken
  by many in Computational Neuroscience, Machine Learning and Artificial Intelligence.
  
  We argue instead that Nature has taken precisely the opposite approach, reverse-engineering in another sense. From the molecular machinery
  inside every cell to the ecosystem of the Earth, evolution exploits the bottom-up, self-organising emergent properties of interacting 
  adaptive dynamical systems to create and preserve structure, synergy, order and information. These systems form a kind of fractal network 
  structure, exchanging energy, material and information.
  
  This is the domain of Complex Systems, a branch of science which has grown in recent years from the Applied Mathematics of Dynamical
  Systems. In this field, each component cannot be modelled using a logical or mathematical clockwork, but instead must be 
  treated as an entity whose future behaviour cannot be predicted, neither exactly, nor even statistically, based on observations. 
  Surprisingly,  mathematical theorems from Dynamical Systems prove that signalling between such systems can allow an agent to both model 
  and control
  a complex system to some extent. The ``computation'' in such an agent requires neither logic nor calculation, as it emerges directly from the
  structure and dynamics of interacting components within the agent itself.

  Computational Neuroscience has for decades investigated the role of dynamical systems and chaos
  in the brain, but most researchers have treated complex dynamics as an inconvenient obstacle to understanding, 
  rather than a key aspect of neural circuit function. In addition, the field of Dynamical Systems and Chaos has
  only recently become a major subfield of Applied Mathematics (due mainly to the advent of high-performance
  computing platforms), so the key findings of the field - even when applied to neural circuits \citep{PhysRevLett.64.821} - 
  have not been made known to the bulk of workers in
  Computational Neuroscience. As a result, important functions of neocortex involving their ability to model dynamics
  in the world have been overlooked or neglected.

  This paper proposes a new view of neocortex as a Universal Modeller of Dynamical Systems, in which each region
  learns to lock onto the dynamics of its inputs, characterise and represent their underlying causal parameters 
  (along with their nonstationarity), and act on them through both feedback and behaviour. A detailed role is described for 
  subpopulations of neurons in all cortical layers in each region, along with the semantics of both well-predicted and unpredicted
  activity in subpopulations. We present a computational model for this based on extensions of the Cortical Learning Algorithm 
  proposed by Hawkins' Hierarchical Temporal Memory \citep{HTMWhitePaper}, and a mathematical vector-based description of the model. 

We develop a multi-layer
model of a region of cortex which extends HTM's Cortical Learning Algorithm (CLA) to all 6 layers of neocortex.
Interlayer and inter-region communication by progressive encoding of representation-prediction 
anomalies is proposed as the key factor in generating self-stabilising dynamics of representation and behaviour.

{\it Note:} The ideas proposed in this paper are extensions and generalisations of the current theory of HTM as proposed 
by Hawkins
and Numenta, which treats evolutions of temporally-varying systems as streams of snapshots of spatial data. The key novelty
in this extension is that the temporal dynamics are at least as important as the sequential structure of successive 
spatial snapshots, and that this gives an agent the opportunity to exploit the crucial information-theoretic 
results of treating sensorimotor inputs as signals from a dynamical system (ie one whose underlying dynamics are
compactly represented using mathematical update rules), rather than observations of a system
where one can only learn sequences by rote (ie where sequences can only be recorded). The dynamical version of HTM described
here is no different from the original where the signal represents non-dynamical or arbitrary sequence inputs.

\section{Outline}

\subsection{Hypothesis: The Brain as a Universal Dynamical Systems Modeller}
We begin with the overall hypothesis of the paper, which is that the neocortex is a hierarchy-like network of interconnected
and coupled regions, each of which is operating as a dynamical system modelling a dynamical system. Using results from the Applied
Mathematics field of Dynamical Systems and Chaos, we will show that this modelling is feasible, simple to implement and applies
pervasively in all kinds of natural and social contexts. In addition, the coupled dynamical systems approach has been shown to
be the only feasible approach to many real-world problems which cannot be tractably solved analytically, and we argue that evolution
discovered this fact millions of years ago and exploited it in neural systems, and in a particularly structured way in neocortex.

\subsection{Overview of Relevant Results in Dynamical Systems}
We introduce a number of concepts and results from Dynamical Systems theory, which provides the mathematical and information-theoretic foundation
for the paper's hypothesis. In particular, we review a number of findings about the power of synchronised and reconstructed dynamical systems
in extracting information of use to a computational agent.

\subsection{HTM as a Universal Dynamical Systems Modeller (UDSM)}

We show how HTM provides a solid computational foundation for the modelling of dynamical systems, and present a multi-layer extension of the
Cortical Learning Algorithm which
describes the role of populations of neurons in all layers of cortex.

\subsection{Cognitive Applications of a HTM UDSM}
A number of cognitive functions can be viewed as involving the controlled coupling of dynamical systems. We describe a number of 
such functions, including motor control in animals, language in humans, and social interaction. We speculate that this literal
interpretation of cortex as a Universal Dynamical Systems Modeller will be useful in understanding and even treating a number of
conditions, including dyslexia and schizophrenia.

\section{Hypothesis: The Neocortex is a Universal Modeller of Dynamical Systems}
The field of Applied Mathematics was transformed in the 1960s and '70s by dramatic discoveries in Dynamical Systems and Chaos. Long
avoided by mathematicians over a century after Poincar\'{e}, major progress became possible using newly available computer hardware, 
software and methods. In the late '70s and early '80s, important mathematical results by \cite{takens1981detecting} and others 
showed that the dynamics of many real-world dynamical systems could be directly and accurately modelled based only on simply
observing measurements of the real-world systems over time. This mathematical property allows an agent to characterise an observed
system and perform short-range forecasting of its future behaviour, without having access to the underlying causal mechanics or 
latent parameters of the system.

Our hypothesis is that the structure of a region of neocortex has evolved to use a time series of sensory (and sensorimotor) inputs
to generate a dynamical analogue of the system generating the sensory inputs, to forecast its short-term future, to identify
stable and slowly-changing characteristics of the system which indicate its hidden controlling parameters or state, to model the 
nonstationarity of the system state, and to interact with the system using time series of motor outputs. 

\section{Dynamical Systems and Chaos} \label{dynamicalsys}

A Dynamical System is a mathematical model whose dynamics are characterised by express update rules, typically differential equations 
in continuous systems, or difference equations in discrete time (a comprehensive survey of Dynamical Systems is \cite{strogatz2014nonlinear}).
Study of such systems began with the advent of calculus, and many simple systems have been studied. Simple Harmonic Motion (SHM), 
an idealised approximation of small oscillations of a simple pendulum or spring, is the archetypical dynamical system introduced in
high school physics. Such systems can be solved exactly using calculus, and most systems studied by engineers and scientists are similarly
straightforward to understand and reason about. However, outside the boundaries of approximation used for SHM, even simple springs
are no longer subject to perfect analysis, as their dynamics are nonlinear and can even be chaotic. 

For these reasons, Engineering and Science have historically avoided the problem of analytically insoluble dynamical systems, usually
by approximating the real system by something akin to SHM, a process called linear approximation. With the advent of computers in the mid-20th
century, however, researchers have been able to study the dynamics of nonlinear and complex systems, and since the 1960s this has become a 
primary focus of Applied Mathematics. 

In the seminal study which triggered the revolution in Dynamical Systems, 
\cite{lorenz1963} described a simple model of atmospheric convection, using three coupled differential equations:

\begin{align}
\frac{\mathrm{d}x}{\mathrm{d}t} &= \sigma (y - x), \\
\frac{\mathrm{d}y}{\mathrm{d}t} &= x (\rho - z) - y, \\
\frac{\mathrm{d}z}{\mathrm{d}t} &= x y - \beta z.
\end{align} 

\begin{figure}[H]
 \centering
 \includegraphics[scale=0.235,keepaspectratio=true]{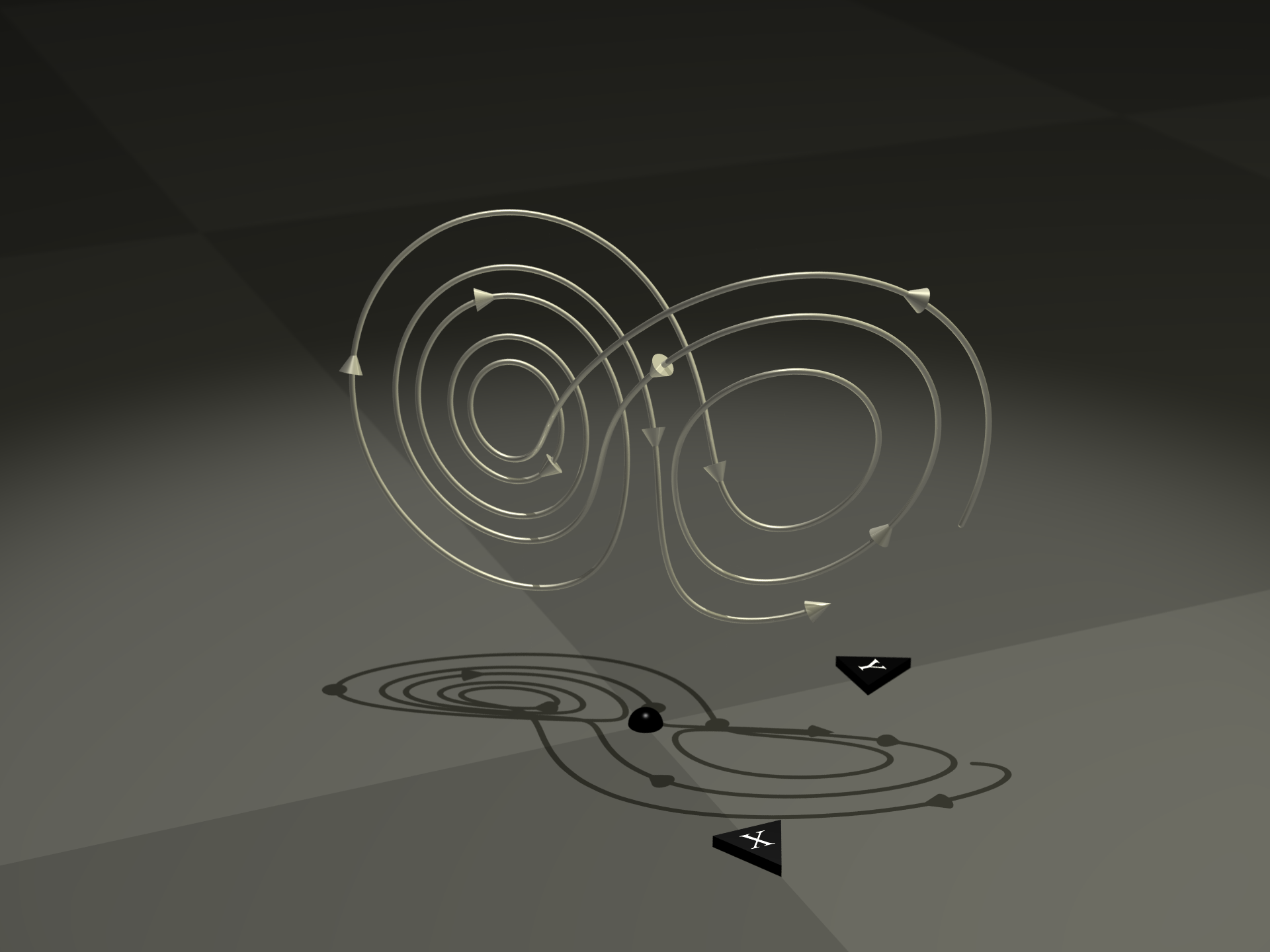}
 \caption{A visualisation of Finite segment of a trajectory of Lorenz's equations, 
 computed by numerical integration and rendered as a metal wire. Parameter values are classical. 
 Conical beads indicate the direction of travel. The black hemisphere marks the origin, and directions 
 of the x and y axes are as indicated. The shadow is cast by a spotlight high on the z axis. \citep{wikiLorenz}}.
 \label{fig:lorenz}
\end{figure}

The quantities $x$, $y$ and $z$ are abstractions of the state of the convecting fluid (here $x$ is proprtional
to the intensity of the convective motion, $y$ to the temperature difference of the ascending and descending currents,
and $z$ to the departure of the vertical temperature profile from linearity). While abstract, these quantities exactly
describe the state of motion of the entire system as a function of time. We can visualise the system by drawing plots of
the trajectories of points $(x(t),y(t),z(t))$ in the {\it phase space} of the system (see Figure \ref{fig:lorenz}). 

The deterministic evolution of the Lorenz system depends on the choices of $\sigma$, $\rho$ and $\beta$, which are together
known as the system's {\it control parameters}. Lorenz studied the system with $\sigma = 10$, $\rho =  8/3$ and $\beta = 28$,
which places the system in a {\it chaotic} regime, in which only short-term forecasts of the state are possible, and the exact 
trajectory of the system exhibits {\it sensitivity to initial conditions}.

\begin{figure}[H]
 \centering
 \includegraphics[scale=0.9,keepaspectratio=true]{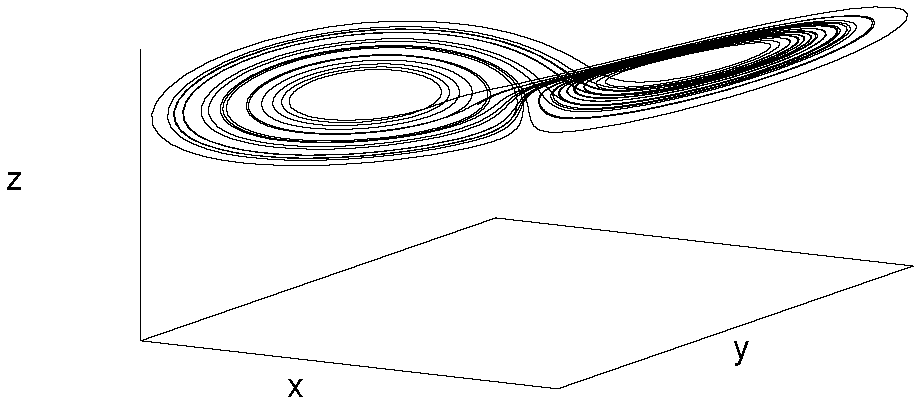}
 \includegraphics[scale=0.9,keepaspectratio=true]{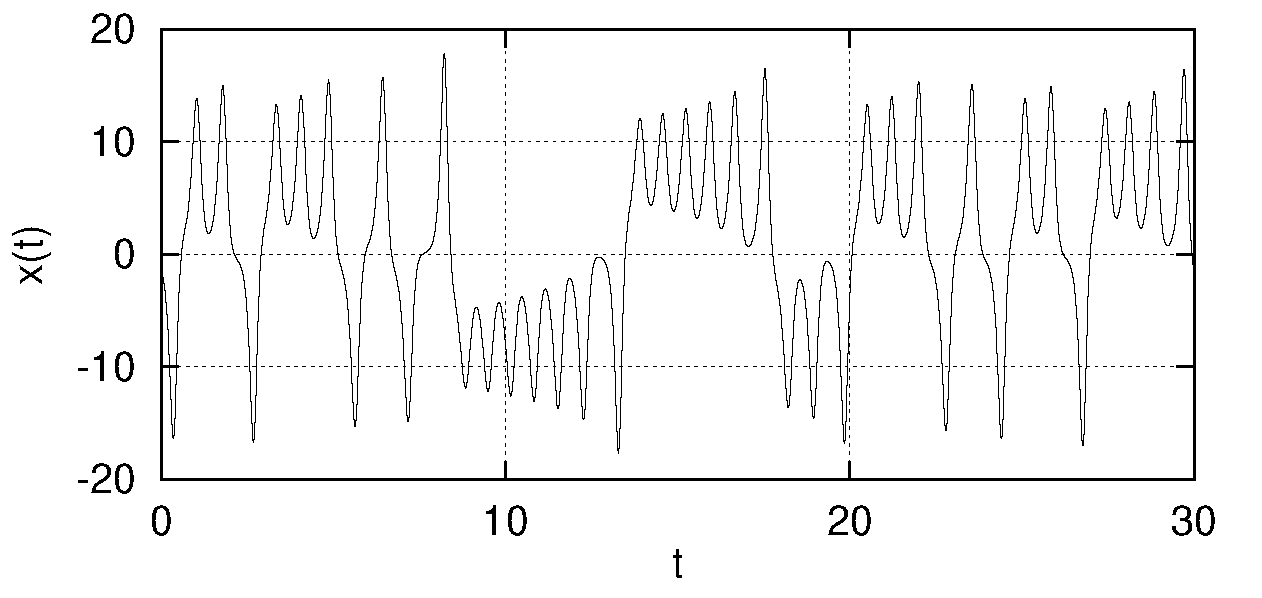}
 \includegraphics[scale=0.9,keepaspectratio=true]{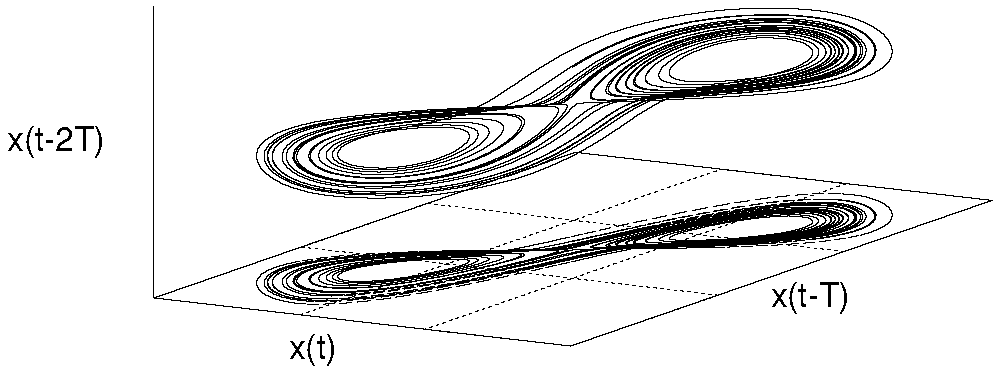}
 \caption{Attractor Reconstruction of the Lorenz Attractor: (top) the Lorenz Attractor in 3D; (centre) a plot of the 
 observed time series $x(t)$; (bottom) reconstruction of the attractor by plotting $\left(x(t),x(t-\tau),x(t-2\tau)\right)$. 
 From \cite{Sauer:2006}}
 \label{fig:reconstruct}
\end{figure}

Floris \cite{takens1981detecting}, basing his study on results going back to \cite{whitney1936differentiable}, proved that
a system such as Lorenz' could be reconstructed in all important details using only a time-series of a single measurement from the
system. A system with manifold dimension $m$ can be reconstructed simply by plotting, for each time $t$, the vector of time-delayed
measurements $\left(x(t),x(t-\tau),x(t-2\tau),...x(t-k\tau)\right)$ of $k>2m+1$ observed values $x$. \cite{Sauer:2006} showed that 
Takens' Theorem holds in less restricted systems, for either $k$ separate measurements at time $t$ or for a set of $k$ time-delayed 
measurements (see Figure \ref{fig:reconstruct} - the bottom shows
a reconstruction of the Lorenz system at top).

As can be seen in the figure, the reconstruction is not visually identical to the original system. It is, however, topologically identical
in an important sense called {\it diffeomorphism}. This means that key analytic properties of the observed system, in particular its
divergence characteristics, are replicated in the reconstruction. Predictions made using the replica have the same properties as the
near-term future of the observed system.

This fact about well-behaved systems such as Lorenz' has been empirically observed to hold in all kinds of natural and real-world
phenomena, even in systems where Takens' Theorem cannot be shown to stricly hold. We claim that Nature has taken advantage of this
fact about reconstruction and synchronization of complex latent dynamics in neocortex, in order to perform computations about the 
real world without access to the underlying mathematics or machinery for solving the differential equations.

An agent exploiting this aspect of Dynamical Systems Theory can perform the following functions:
\begin{enumerate}
  \item using very simple computations on the signal, the agent can generate a simulacrum of the observed system which preserves
  the essential properties of that system;
  \item by modelling or learning the transitions between points in the reconstructed space (alternatively the vector field in the tangent
  space of a point), the agent can run a simulation forward in time to perform forecasting;
  \item by making multiple predictions of the transition vectors, and comparing the resulting predictions with actual trajectories,
  the agent can remain locked on to a nonstationary system;
  \item by monitoring and tracking the correct predictions of transitions, the agent can characterise the controlling parameters (such 
  as $\sigma$, $\rho$ and $\beta$ in Lorenz' system), and track those parameters over time if the system is nonstationary.
  \item by combining prediction with measurement, the system can distinguish between the chaotic signal and 
  non-chaotic
  inputs - including noise and any signal - using the chaos as a carrier signal.
  \item one agent can communicate arbitrary information to another by using such coupled dynamical systems; see \cite{Marcus:2008} for an 
  overview of
  Symbolic Dynamics.
  \item an agent can control a dynamical system such as an arm, leg, an entire body, or a body-tool assemblage by sending controlling
  signals to the driving inputs of that system; see \cite{Ott:2006} for a guide to this.
\end{enumerate}

We will describe a subset of these functions.

\subsection{Reconstruction of an Observed System} \label{reconstruction}
Real-world systems found in nature, in society, and in economics are difficult or impossible to study using the standard 
tools used for centuries since Newton and Leibniz. In most cases, we have access neither to the internal dynamics (the differential
or difference equations governing the update rules), the control parameters which vary the global properties of the system, nor 
the exact values of the observables of the system. Engineers typically address these problems by modelling the system approximately
using simpler, linear dynamics and tools for estimating their governing parameters and measurement error.

Several decades of research into the information theory and mathematics of such systems has shown that there is a simpler and
in many ways more powerful method: {\it reconstruction} of the dynamics using signals from the observed system \citep{takens1981detecting,Sauer:2006}.
The essence of the strategy is that sufficient information about the observed system is encoded in its signal to allow its reconstruction
in an observing system using very simple computations.  

A classical Dynamical System such as Lorenz' can be seen as a {\it vector field} or {\it flow} in its phase space, where the value of the
vector field at each point is effectively a velocity vector indicating how a ``particle'' at the point will move, tracing out a trajectory over
time (see \ref{fig:lorenz}). When the system is chaotic, no long-term forecast of its position can be made, because nearby trajectories diverge 
exponentially quickly as a function of time. However, for sufficiently short timescales, nearby trajectories can be shown to remain within
a required volume. This means that, despite uncertainty due to measurement noise (or noise inherent in the system), the near-future evolution
of the system can always be predicted within finite bounds.

The key result of Takens' Theorem is that this property is preserved in the reconstructed system (in fact this is the main practical 
use of the theorem). Formally, the two systems are {\it diffeomorphic}, which simply means that they have the same topological properties.

Informally, the key insight is that the reconstructed system is to all intents and purposes {\it the same thing} as the observed one,
but represented in a form fully accessible to the observer, without needing a mathematical model of the original dynamics of the real
system. 

\subsection{Learning to Predict using Reconstruction}

An agent can record a long time series of observed signals from a Dynamical System using this method, recording for each point in the 
reconstructed space the transition vector observed from that point to its successor. Over time, this allows the agent to estimate the flow 
field in the neighbourhoods of more and more observed points in the signal. If the vector field is sufficiently smooth, this allows the 
agent to estimate the successor of a new data point by combining (in some simple way) the transition vectors of its near-neighbours in 
the recording.

\begin{figure}[H]
 \centering
 \includegraphics[scale=0.68,keepaspectratio=true]{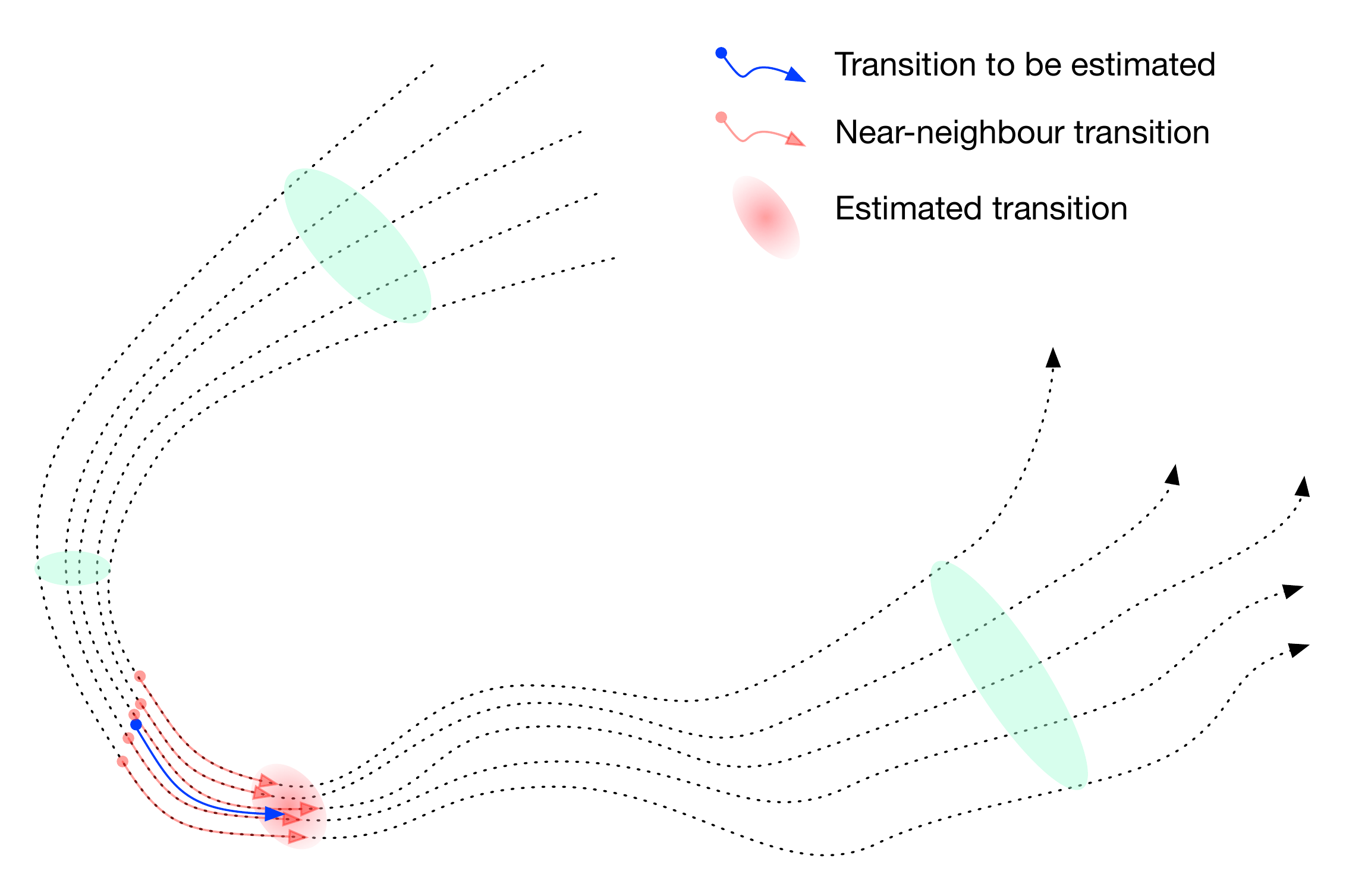}
 \caption{Prediction of future position using near-neighbour transitions. Five trajectories in recontructed phase space are shown, 
 which get close together at bottom left. 
 A new data point (blue dot) is presented, and we choose the closest points on the nearby trajectories (red dots) to form an estimate (red cloud)
 of the future position of the new input.}
 \label{fig:prediction}
\end{figure}

We have described a somewhat na\"{i}ve method for prediction of a transition, which involves recording every transition at every point in
the past, searching over them for every new point, choosing some number of near neighbours, and calculating the estimate from the chosen
transition vectors. This procedure can be dramatically optimised using appropriate machine learning methods, as we shall demonstrate using
HTM.

\subsection{Multiple Predictions and Non-stationarity}

In real-world systems, underlying control parameters may change over time, a condition known as {\it non-stationarity}. For example, as we grow 
our bones change length, our muscles strengthen, and our weight changes. This changes the dynamics of the resulting system from one {\it regime}
to another. An agent may discover and characterise this non-stationarity by making multiple predictions, and observing which of them is fulfilled.

\begin{figure}[H]
 \centering
 \includegraphics[scale=0.92,keepaspectratio=true]{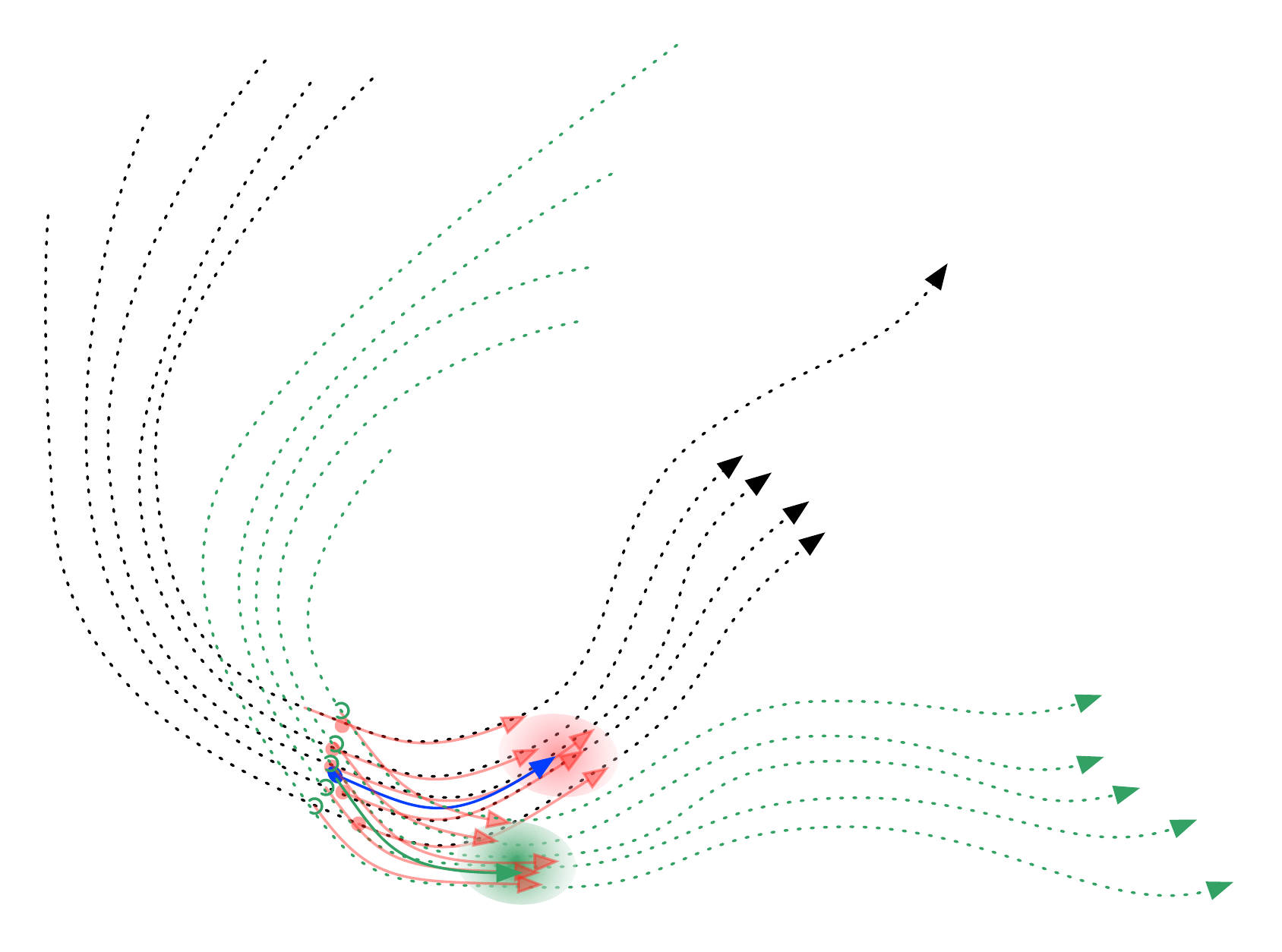}
 \caption{Multiple Predictions. Trajectories for two slightly different dynamical systems (black and green dashed lines). Given a new data 
 point (blue dot), the agent can make two distinct predictions (red and green clouds), by using near-neighbours from both classes (red and 
 green dots). The system may be non-stationary, ie the control parameters may have shifted from the black regime to the green, in which
 case the next input will distinguish between regimes.}.
 \label{fig:multiprediction}
\end{figure}

\subsection{Prediction Error and Correction Vectors} \label{predictionError}

Looking again at the multiple prediction visualisation (Figure \ref{fig:multiprediction}), consider a situation where the agent, unsure of which
regime is in play (which neighbours to trust), chooses to predict {\it both} red and green clouds, and the next input is in fact the green one. 
In this case, we could consider the green arrows as {\it correctly predicting} and the red as incorrect. In this case, we can view the vector 
from blue to green estimates as a {\it prediction error vector} or alternatively as a {\it correction vector}. This information can be used by
a learning system to make better predictions in future (if the green regime remains in place), to assist in classifying the regime, to adjust the
influence of the neighbours, and so on. We shall see that this aspect of the dynamical modelling is important in the HTM-based system.

\subsection{Driven Dynamical Systems}
Reconstruction of dynamical systems is a very interesting and useful concept for working with complex dynamical systems.

A related but crucial property of Dynamical Systems is where a signal from one is used as an input to the controlling
equations of the other, a process known as {\it driving}. The dynamics of the driven system are altered in a way which merges
information about the driving system with its own states and dynamics. 

This process is the key to interaction between agents and their environments. Sensory information is a signal which drives the dynamics of
primary sensory cortex, interacting in turn in complex ways with other parts of cortex, and also with the dynamical systems of the body
via motor signals.

\subsection{Multiple Systems, Synchronization and Causality}
These ideas can be extended to model multiple systems, allowing the agent to determine and detect the nature of the relationships
between different objects, systems of objects, and sensorimotor modalities of single systems. The study of the relationships between 
different systems involves the concept of {\it synchronization}, defined in \cite{brown2000unifying} as follows:

``Two dynamical systems $\mathbf{x}$ and $\mathbf{y}$ are {\it synchronized}
with respect to the properties $g_x$ and $g_y$ if there is a time
independent mapping $\mathbf{h}:\mathbb{R}^k \times \mathbb{R}^k \rightarrow \mathbb{R}^k$
such that
$\lVert{\mathbf{h}(\mathbf{g}(\mathbf{x}),\mathbf{g}(\mathbf{y}))}\rVert = 0$
holds on all trajectories.''

An alternative definition from \cite{pikovsky2003synchronization} is ``the appearance if a statistical relationship between the observables 
of two systems due to mutual interaction''.

Note that in the case of an observed system and its reconstruction, there is a trivial synchronization whereby the observed signal and the
reconstructed vector of lagged signal values are connected directly in a {\it master-slave} synchronization.

The most interesting kind of synchronisation is {\it Generalised Synchronization}, defined originally in \cite{rulkov1995generalized}, which
involves a functional relationship between the variables of $\mathbf{x}$ and $\mathbf{y}$, ie. $\mathbf{x}(t) = \phi(\mathbf{y}(t))$.

Attempts to use reconstruction to characterise the relationships between putatively synchronized dynamical systems have had limited success 
\citep{kato2013limits}, 
but recent progress has been made using a number of relatively simple innovations. In particular, \cite{chicharro2009reliable} and 
\cite{sugihara2012detecting} have developed
methods which use ranking of near-neighbours in phase space to overcome difficulties in handling noise, tranposing between systems of different
dimension, and normalising the volumes occupied by neighbourhoods in each system. Chicharro and Andrzejak define their L-index by:

$$L(X|Y) = \sum\limits_{i=1}^N \frac{G_i(X)-G_i^k(X|Y)}{G_i(X)-G_i^k(X)}$$

where $G_i(X) = \frac{N}{2}$ and $G_i^k(X) = \frac{k+1}{2}$ are the mean and minimal mean rank, respectively, for $k$ nearest 
neighbours and $N$ data points. $G_i^k(X|Y)$ is the mean rank of $k$ neighbours of $X_i$ using the $k$ neighbours of $Y_i$ to
choose indices. The index is thus a dimensionless measure of the volume increase of a neighbourhood in $X$ when using points
from $Y$ to define neighbours. The authors demonstrate the robustness of this method for both artificial (Lorenz) and natural
systems.

Interestingly, this use of ranking is also a property of
the activation-ordered formation of Sparse Distributed Representations in HTM (see \cite{byrne2015paCLA} and Section \ref{htm-as-udsm}).

\begin{figure}[H]
 \centering
 \includegraphics[scale=0.85,keepaspectratio=true]{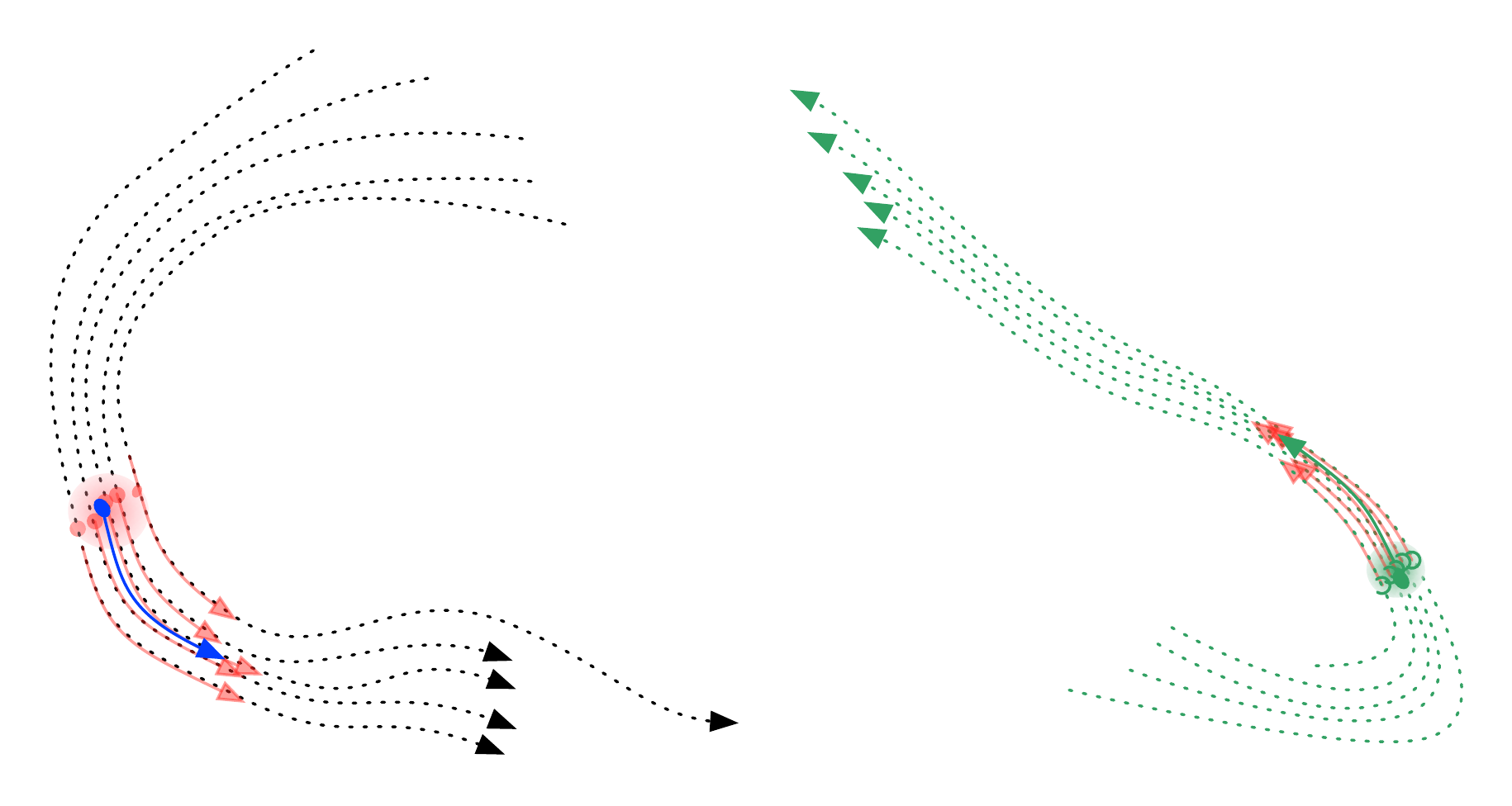}
 \caption{Convergent Cross Mapping \citep{sugihara2012detecting} and L-index \citep{chicharro2009reliable}. If
 the system $X$ at left is driven by $Y$ at right, then a neighbourhood of a point $Y_i$ can be used to identify
 candidate points in $X$ which will occupy a volume around $X_i$.}.
 \label{fig:ccm}
\end{figure}

\newpage
\section{HTM as a Universal Dynamical Systems Modeller (UDSM)} \label{htm-as-udsm}

Dynamical Sysems Modelling using classical point-neuron networks (usually known as Artificial Neural Networks or ANNs) has been studied
in the past (see \cite{masulli1999neural} for example), but in a feedforward context where the ANN is used only to read out the 
state, forecast evolution, and anomaly profile of the observed system. In these models, the ANN is not itself a dynamical system in a 
complex network.

Here, we develop a multi-layer model for HTM which describes how a region of neocortex learns to model and interact with
dynamical systems, whether external or internal to the agent. 

\subsection{Prediction-Assisted Cortical Learning Algorithm - Review}
Hierarchical Temporal Memory and the Prediction-Assisted Cortical Learning Algorithm (paCLA) are described in considerable 
detail in \cite{byrne2015paCLA}, as an extension of HTM theory as described most recently in \cite{hawkins2015neurons}.
In essence, a series of sensorimotor inputs, each represented as a semantically-based sparse distributed representation
vector, is projected into a layer's output space of cell activations, forming an SDR representing each input. Pattern Memory is the 
learning (via synaptic adaptation in proximal dendrites) of this spatial mapping. In addition, the layer uses recurrent distal 
connections to learn and predict transitions between output SDRs, a process called Transition Memory. The output of the layer
is a combination of orthogonal vectors, one formed from highly-predictive cells, the other signifying the difference between prediction
and reality.

paCLA models cortical Layer 4 (L4) as just described. A second layer (collectively called L2/3) can use this L4 output to learn to form a more stable representation 
of a sequence or cycle of lower-layer SDRs, a process called Temporal Pooling. L2/3 outputs a series or sequence of SDRs which 
characterises the succession of higher-level meta-states in the input stream. This output is passed up the hierarchy for further 
processing. We now extend this simple, feedforward sensory processing model to an entire region of 6 layers. We begin by connecting the form of 
signals in HTM, the Sparse Distrbuted Representation, with the concepts from the section on Dynamical Systems.

\subsection{Sparse Distributed Representation as High-Dimensional Embedding}

We've seen from the mathematics of \cite{takens1981detecting}, \cite{Sauer:2006} and \cite{whitney1936differentiable}
that a reconstruction or synchronization of dynamical systems can be achieved when sufficient spatial and/or temporal
dimensionality is represented by an agent using the observed signal. This is certainly true in HTM when considering the 
representational capacity of SDRs of typical sizes (see \cite{SDRpaper}). As mentioned in \cite{byrne2015paCLA}, 
k-Winner-Take-All (kWTA) representations are highly efficient universal function approximators (via \cite{WTAMaass2000}),
and \cite{chicharro2009reliable} identified the kWTA near-neighbour representation as optimal for automatic
analysis of the causality relationships of putatively coupled dynamical systems. SDRs clearly unite these factors of 
dimensionality and efficient kWTA functional representation.

\subsection{Multilayer Flow Model of HTM}

\begin{figure}[H]
 \centering
 \includegraphics[scale=1.3,keepaspectratio=true]{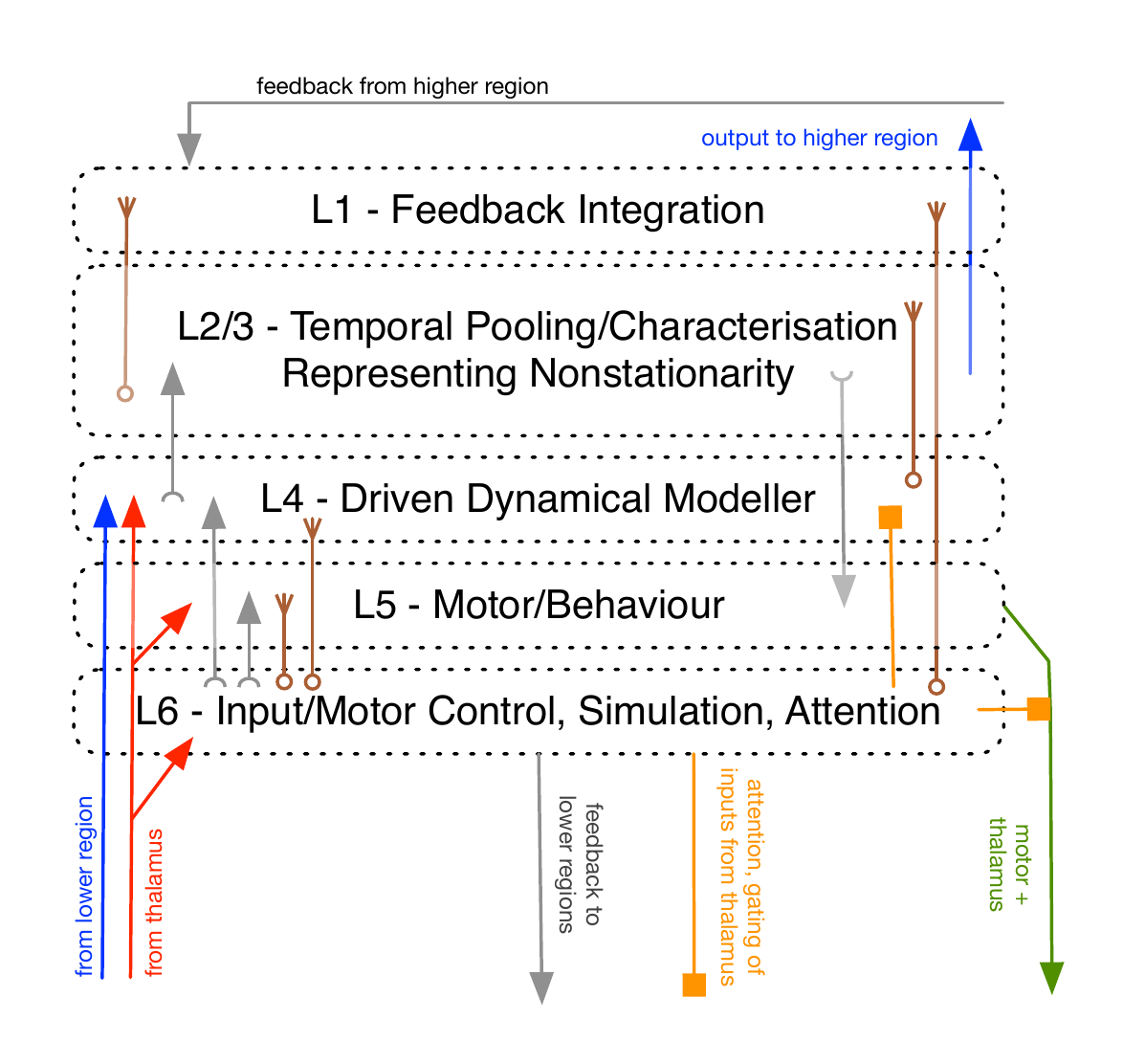}
 \caption{Flows of inputs and control signals in Multilayer HTM. Sensory (red) inputs from thalamus and (blue) from lower regions flow to L4 and 
 L6. L4 models inputs and transitions, L2/3 temporally pools over L4, passing its outputs up. L5 integrates top-down feedback, L2/3 outputs, and 
 L6 control to produce motor output. L6 uses gating signals (orange) to co-ordinate and control inputs and outputs, and can execute simulation 
 to run dynamics forward.}.
 \label{fig:multilayerFlow}
\end{figure}

A region of neocortex is itself a network of dynamical circuit systems, with flows of signals entering from sensory and motor sources,
from other regions of cortex in the network, from subcortical structures, and from relay nuclei in the thalamus. Within the region, a
specific network of connections directs the interactions between cell populations in the various layers. The content, sparseness and dynamics
of these signals dynamically interact with the inherent learned dynamics of each circuit, and gating circuits control whether and how 
signals travel in the network. The outputs of the region include signals up and down the hierarchy and into the cortical network, 
gating signals controlling information flow in and out of the region, interaction with subcortical structures, and motor behaviour.

\subsection{Layer 4 as a Transition Memory Modelling Input Dynamics}

Layer 4 of neocortex is the primary recipient of afferent sensorimotor inputs to neocortex, as well as the main recipient 
of corticocortical ``feedforward'' signals in hierarchy. We claim that the purpose of Layer 4 is to represent its input in a
high-dimensional form, in order to reconstruct the dynamics of its inputs. Using the Prediction-Assisted Cortical Learning 
Algorithm (paCLA) described in \cite{byrne2015paCLA}, L4 can represent each input as a high-dimensional vector, and can learn 
the transition vectors predicted by the history of inputs to the layer. 

Importantly, a layer of cortex using paCLA represents each input as a sorted sequence of action potentials, ranked by 
activation arising from a combination of predictive and actual sensory inputs, in a fashion ideally modelled using the mathematics of 
\cite{chicharro2009reliable}. This allows L4 to represent its perception as a combination of predicted dynamics and a 
correction vector of bursting columns, as illustrated in Subsection \ref{predictionError}.

Layer 4 is effectively ``plotting'' its inputs in a high-dimensional space of SDRs, thereby fulfilling the first key 
representational requirements of the mathematics of reconstruction of dynamical systems. Secondly, L4 learns to predict 
the future evolution of its inputs using the transition vectors represented by its distal dendrites. It thus models both
the instantaneous state (position in phase space) and the vector field representing its future trajectory.

\subsection{Layer 2/3 Temporal Pooling Models Control Parameters and Nonstationarity}

As described in \cite{byrne2015paCLA}, a subpopulation of neurons in Layer 2/3 of cortex performs a function known as 
Temporal Pooling, representing sets of successively predicted SDRs in Layer 4 as a single, stable output. This TP SDR 
can be seen as a kind of dynamical symbol for the regime currently experienced by L4, which encodes the particular
latent control parameters of the system L4 is modelling. 

It's important to note that Temporal Pooling is itself a dynamical phenomenon, depending as it does on the extent to which
the L4 neurons can predict transitions in their inputs. If the observed system is nonstationary, the transitions in L4 will
shift to track new trajectories on altered attractors, represented by different neurons in L4, and so the L2/3 representation 
will itself transition to represent the new regime. Any prediction errors in L4 are passed to L2/3 separately, causing L2/3
to partially cease pooling over the old transitions and gradually begin pooling over new ones. This can itself be viewed
as shifting in L2/3 of its SDR from one basin of attraction to another.

Layer 2/3 can learn sequences of such stable, temporally pooled representations, and incorporate contextual and top-down
feedback to model the higher-level structure of the system or environment being observed.

This dynamical perspective of L4-L2/3 function operates in conjunction with the more sequence-centric and chunk-centric view
originally proposed for Temporal Pooling by \cite{TPNewIdeas}. Which function predominates depends on the kind of inputs being 
processed by the region. 

\subsection{Layer 5 Dynamically Models Motor Interactions between the Region and the World}

The primary function of Layer 5 is to produce behaviour/motor output, via signals sent to subcortical motor centres and other parts of the brain. 
This happens in every region of cortex, not just in so-called motor cortex (which is primarily dedicated to large limb movements). The reason is 
that each region can use local information to interact directly with its source of sensory information - the system it's modelling.

Layer 5 is the main integration layer in neocortex. Its primary pyramidal neurons are the largest in the region, with an active apical 
trunk passing up through Layers 4, 3 and 2, apical dendrites in Layer 1, and basal dendrites in Layers 5 and 6. A L5 neuron thus has access to
all sources of information on the dynamical states of the region, and its purpose is to use this information to affect the system of interest
to the region by sending motor signals.

L5 neurons project their axons to synapse onto cells in subcortical motor centres. This is guided initially by genetic control (to send the 
axons to the correct target area), but then the precise alignment of individual axons to target motor centre cells is based on activity
matching. For example, primary visual cortex (V1) neurons target the Superior Colliculus, a centre using retinal inputs to control
eye movements \citep{triplett2009retinal}.

In addition, L5 axons split in two as they leave the region. The second branch ascends hierarchy in the cortex and informs higher
regions of the motor/behavioural decisions made by a region. This signal complements that generated by L2/3, giving other areas of the 
brain a full sensorimotor picture of this region's dynamics.

L5 cells learn to form the longest link in the feedback signalling loop between the brain and the world. This is perhaps why it has 
the broadest access both to fast-changing sensory signals and every layer in its region. 

\subsection{Layer 6 Coordinates, Controls and Simulates Signalling and Activity} 

Layer 6 is the most complex of all cortical layers. Its numerous interconnected circuits together perform key ``operating system-like'' functions
which coordinate and control the activity, communications and dynamics of the region:

\begin{enumerate}
  \item L6 processes the same inputs as L4, but uses extra information from L5 (and thus from L1-L2/3-L4) to control its dynamics.
  \item L6 uses projections to thalamus to gate inputs to the region, deciding how much ``reality'' the region sees, and controlling attention.
  \item L6 projects strongly to L4, allowing it to provide artificial inputs to the region for forward simulation.
  \item L6 projects feedback signals to lower regions, providing their L2/3 with top-down context, assisting lower-level sensory interpretation.
  \item Feedback to lower regions from L6 also provide top-down signals controlling finer-grained L5 motor output in lower regions (unfolding 
  motor sequences)
  \item L6 gates the activity and motor output of L5, providing control over whether actions really happen.  
  \item Some L6 circuits form a dynamical ``control system'' which interacts with and controls how each of the above systems evolve.
\end{enumerate}

Combining all these circuits, it is clear that L6 is the conductor of operation of the region, deciding dynamically what region-level 
functions are carried out, coordinating with other regions to perform a larger-scale task. 

\subsection{Learning and Meaning in Dynamical HTM}
While the above is a plausible model for neocortex, it somewhat begs the question of how this highly complex fractal network structure
of interacting dynamical systems can be made to work. The answer is actually quite simple.

In early development, the gross circuit layout described above is established using well-understood genetic mechanisms. The formation of
specific cellular connections and synapses between neurons takes place, however, through activity-based learning 
(as described for V1 neurons in \cite{triplett2009retinal}).  

Motor and sensory modelling is learned through activity and behaviour, in a process described by \cite{warren2006dynamics} and 
\cite{freeman2002limbic} as the Action-Perception cycle. 

\cite{freeman2003meaning, freeman2004creatingmeaning} describes meaning in neural systems as constructed from the interaction of
incoming information with the dynamics of cortex (itself the learned product of its history of interaction with the world).

\subsection{Implementing Dynamical HTM}
A system for demonstrating DHTM and experimenting with its properties is in the planning stage at time of writing. A single layer implementing
the Prediction-Assisted Cortical Learning Algorithm (paCLA) is the basic building block for DHTM, as it has the required 
properties \citep{byrne2015paCLA}. The outstanding issue concerns the design and configuration of the connections between layers.

Comportex \citep{felix_andrews_2015_34602} is an experimental platform for developing HTM algorithms, and includes a version
of paCLA (differing in having a single proximal dendrite per column rather than per cell). NeoRL \citep{222464_2015_35283} also uses aspects of
paCLA in a reinforcement learning setting. Clortex \citep{fergal_byrne_2015_34545} is a system-level design which has properties useful 
for large-scale deployments.

\newpage
\section{Cognitive Applications of a HTM UDSM}

\cite{branicky1995universal} and others have proven that a number of discrete and hybrid dynamical system designs can simulate any Turing 
Machine, and thus possess the power of Universal Computation. As stated earlier, \cite{maass1997networks} has shown that k-Winner-Take-All binary
networks are also capable of replicating any binary function. While theoretically satisfying (similar proofs are used to justify study of 
recurrent neural networks), this does not tell us how such networks can be efficiently constructed to solve a given computational task. 

Deep Learning uses large amounts of labelled data and powerful computational resources to exhaustively train very large and deep networks
of simple, identical point neurons. Given a domain, with training data of sufficient size and quality, such networks can do very well. They can
even appear to mimic in detail an entire pipeline of the brain's sensory cortex \citep{NIPS2013_4991}. 

So, since standard computer programs, ANNs and HTM-based systems can all be seen as equivalently powerful, the question of which
mechanism is suitable as a substrate for intelligence is about how the programming of such systems can be practically achieved. 

While computer software might be considered optimal in terms of efficiency of execution, compactness of representation, and comprehensibility,
it is implausibly expensive to hand-code all possibly relevant program modules, and then the difficulty becomes a combinatorially explosive
search for the right choice of software module in a given situation (see \cite{Hutter:04uaibook}).

Deep Learning Networks of various kinds are better for many simple tasks, such as recognising objects in a static image. A DL network can be viewed
as something which gradually sculpts itself using large amounts of correctly prepared data. Recent innovations in Deep Learning have added 
numerous extra capabilities to them, for example by adding gated memory cells \citep{hochreiter1997long}, or by combining them into reinforcement-learning
programs \citep{mnih2013playing} which can quickly learn to play Atari video games. 

While impressive, these hybrid systems often require non-neural components which tend to be very task-specific in order to get any
level of performance. For example, the Atari-playing system samples from a database of all past video frames to counteract the destructive effects
of correlations between recent frames. This is needed because the underlying mathematics of the Deep Learning network requires
randomly-chosen, uncorrelated (IID) samples in order to avoid catastrophic instability. Indeed, this workaround is one of the reasons for the
recognition of the DeepMind paper in 2014 - before that, combining Deep Learning and Reinforcement Learning had been regarded as unlikely to work. 
One would think that recent temporal context would be useful to an agent playing a video game, but this is precluded by the dynamics of 
the network used. 

In addition, learning in most or all Deep Learning relies on some kind of backpropogation of errors using global errors and derivatives of continuous
transformation functions in potentially all the neural units. This creates a number of problems for developers and users, problems which are not 
reflected in mammalian neocortical learning. These include the inability to choose non-differentiable transformation functions; exploding or
vanishing gradients; a need to process a large proportion of connections and neurons; requirements for very large datasets; batch or minibatch 
processing; randomisation of inputs to remove temporal correlations. 

These issues require workarounds which need to be judiciously applied by skilled network architects, and must be tailored for each task and dataset. 
Most importantly, they often result in the discarding of important information in the data, such as temporal coherence.   

We argue that by combining the constraints on structure of paCLA with the power of learned interaction in coupled dynamical systems, regions of
HTM neurons can learn to exploit the dynamical temporal structure of the data and simultaneously gain the ability to efficiently interact with
the internal and external world.

\cite{beer2000dynamical} is a very useful introduction to cognition from a general dynamical systems perspective. \cite{kello2007emergent} provides
convincing evidence of emergent coordination of dynamical processes in human behaviour, arguing that the pervasive appearance of uncorrelated
$1/f$ scaling (characteristic of emergent coordination) across many measures and activities can only be parsimoniously explained in this way.

\paragraph{Motor Control in Animals}

Many researchers have used dynamical systems as a model of skilled motor control.
\cite{taga1991self} present a model of bipedal locomotion based on coupled neural oscillators which are controlled by low-dimensional 
top-down inputs. \cite{ijspeert2002movement} describes a robotic system which learns to execute desired movements (reaching, drawing 2D shapes, 
and tennis swings) using coupled dynamical systems.

\paragraph{Language in Humans}

\cite{elman1995language} described a model of language in terms of dynamical systems. His work can be seen as one of the foundations for recent
breakthroughs by Deep Recurrent Neural Nets in Natural Language Processing and Machine Translation. \cite{webber2015} describes an encoding
system - Semantic Folding - for producing Sparse Distributed Representations for words and documents. This system was directly inspired by 
HTM theory, and work has begun on exploring the properties of the representations when used with HTM systems.

\paragraph{Social Interaction}

One of the first applications of Dynamical Systems theory was in the social sciences, and it is now typical to model
social interactions as communicating dynamical systems in a social network or graph. For an excellent overview, 
see \cite{boccaletti2006complex}.

\paragraph{Disorders from a Dynamical Systems Viewpoint}

We speculate that a number of serious disorders involve dysfunction of, or significant changes to, the processes described here.

\cite{wijnants2012interaction} compared the reading performance of ``dyslexic'' and ``non-dyslexic'' children using methods based 
on Takens' Theorem, and found that reading performance correlated
with the level of self-organisation of the dynamics of interactions between cognitive components 
[Note: ``dyslexic'' refers to reported diagnoses,
see \cite{elliott2008does} for a discussion of this]. \cite{greijn2011dyslexia} further explores this correlation.

\cite{loh2007dynamical} hypothesised a dynamical systems approach to understanding the mechanisms underlying the complex
of disorders in schizophrenia. They modelled this using networks of spiking neurons, and concluded that simple changes to neural
characteristics could cause changes in the dynamics which corresponded to symptoms of the disorder.

\section*{Acknowledgements}

The author wishes to thank Robert Freeman for our discussions about these ideas, for highlighting the need to consider results
from the Applied Mathematics of Dynamical and Complex Systems, and for his presentation of his own ideas on 
language \citep{DBLP:journals/corr/Freeman14}, which motivated much of the research presented here.

I'd also like to thank Richard Crowder (@rcrowder) for his assistance and feedback in preparing this paper.
{\footnotesize
\bibliography{scifile}

\begin{thebibliography}{48}
\providecommand{\natexlab}[1]{#1}
\providecommand{\url}[1]{\texttt{#1}}
\expandafter\ifx\csname urlstyle\endcsname\relax
  \providecommand{\doi}[1]{doi: #1}\else
  \providecommand{\doi}{doi: \begingroup \urlstyle{rm}\Url}\fi

\bibitem[Ahmad and Hawkins(2015)]{SDRpaper}
Subutai Ahmad and Jeff Hawkins.
\newblock {P}roperties of {S}parse {D}istributed {R}epresentations and their
  {A}pplication to {H}ierarchical {T}emporal {M}emory.
\newblock arXiv:1503.07469 [q-bio.NC], Jul 2015.
\newblock URL \url{http://arxiv.org/abs/1503.07469}.

\bibitem[Andrews and Lewis(2015)]{felix_andrews_2015_34602}
Felix Andrews and Marcus Lewis.
\newblock {comportex: Hierarchical Temporal Memory in Clojure}, December 2015.
\newblock URL \url{http://dx.doi.org/10.5281/zenodo.34602}.

\bibitem[Beer(2000)]{beer2000dynamical}
Randall~D Beer.
\newblock Dynamical approaches to cognitive science.
\newblock \emph{Trends in cognitive sciences}, 4\penalty0 (3):\penalty0 91--99,
  2000.

\bibitem[Boccaletti et~al.(2006)Boccaletti, Latora, Moreno, Chavez, and
  Hwang]{boccaletti2006complex}
Stefano Boccaletti, Vito Latora, Yamir Moreno, Martin Chavez, and D-U Hwang.
\newblock {Complex networks: Structure and dynamics}.
\newblock \emph{Physics reports}, 424\penalty0 (4):\penalty0 175--308, 2006.

\bibitem[Branicky(1995)]{branicky1995universal}
Michael~S Branicky.
\newblock Universal computation and other capabilities of hybrid and continuous
  dynamical systems.
\newblock \emph{Theoretical Computer Science}, 138\penalty0 (1):\penalty0
  67--100, 1995.

\bibitem[Brown and Kocarev(2000)]{brown2000unifying}
Reggie Brown and Ljup{\v{c}}o Kocarev.
\newblock A unifying definition of synchronization for dynamical systems.
\newblock \emph{Chaos: An Interdisciplinary Journal of Nonlinear Science},
  10\penalty0 (2):\penalty0 344--349, 2000.

\bibitem[Byrne(2015{\natexlab{a}})]{HTMWhitePaper}
Fergal Byrne.
\newblock {H}ierarchical {T}emporal {M}emory including {HTM} {C}ortical
  {L}earning {A}lgorithms.
\newblock Revision of Hawkins and Ahmad, 2011, Oct 2015{\natexlab{a}}.
\newblock URL \url{http://bit.ly/htm-white-paper}.

\bibitem[Byrne(2015{\natexlab{b}})]{byrne2015paCLA}
Fergal Byrne.
\newblock {E}ncoding {R}eality: {P}rediction-{A}ssisted {C}ortical {L}earning
  {A}lgorithm in {H}ierarchical {T}emporal {M}emory.
\newblock \emph{arXiv preprint arXiv:1509.08255}, Sep 2015{\natexlab{b}}.
\newblock URL \url{http://arxiv.org/abs/1509.08255}.

\bibitem[Byrne(2015{\natexlab{c}})]{fergal_byrne_2015_34545}
Fergal Byrne.
\newblock Clortex: Initial pre-alpha release, December 2015{\natexlab{c}}.
\newblock URL \url{http://dx.doi.org/10.5281/zenodo.34545}.

\bibitem[Chicharro and Andrzejak(2009)]{chicharro2009reliable}
Daniel Chicharro and Ralph~G Andrzejak.
\newblock Reliable detection of directional couplings using rank statistics.
\newblock \emph{Physical Review E}, 80\penalty0 (2):\penalty0 026217, 2009.
\newblock URL \url{http://dx.doi.org/10.1103/PhysRevE.80.026217}.

\bibitem[Elliott and Gibbs(2008)]{elliott2008does}
Julian~G Elliott and Simon Gibbs.
\newblock Does dyslexia exist?
\newblock \emph{Journal of Philosophy of Education}, 42\penalty0
  (3-4):\penalty0 475--491, 2008.

\bibitem[Elman(1995)]{elman1995language}
Jeffrey~L Elman.
\newblock Language as a dynamical system.
\newblock \emph{{Mind as motion: Explorations in the dynamics of cognition}},
  pages 195--223, 1995.

\bibitem[Freeman(2014)]{DBLP:journals/corr/Freeman14}
Robert~John Freeman.
\newblock Parsing using a grammar of word association vectors.
\newblock \emph{CoRR}, abs/1403.2152, 2014.
\newblock URL \url{http://arxiv.org/abs/1403.2152}.

\bibitem[Freeman(2002)]{freeman2002limbic}
Walter~J Freeman.
\newblock The limbic action-perception cycle controlling goal-directed animal
  behavior.
\newblock In \emph{Neural Networks, 2002. IJCNN'02. Proceedings of the 2002
  International Joint Conference on}, volume~3, pages 2249--2254. IEEE, 2002.

\bibitem[Freeman(2003)]{freeman2003meaning}
Walter~J Freeman.
\newblock A neurobiological theory of meaning in perception part i: Information
  and meaning in nonconvergent and nonlocal brain dynamics.
\newblock \emph{International Journal of Bifurcation and Chaos}, 13\penalty0
  (09):\penalty0 2493--2511, 2003.

\bibitem[Freeman(2004)]{freeman2004creatingmeaning}
Walter~J Freeman.
\newblock How and why brains create meaning from sensory information.
\newblock \emph{International journal of bifurcation and chaos}, 14\penalty0
  (02):\penalty0 515--530, 2004.

\bibitem[Greijn(2011)]{greijn2011dyslexia}
Lieke~T Greijn.
\newblock \emph{Why dyslexia appears as it does: The view of interaction
  dominant dynamics on the cognitive deficit of dyslexia}.
\newblock PhD thesis, Radboud University Nijmegen, 2011.

\bibitem[Hawkins(2014)]{TPNewIdeas}
Jeff Hawkins.
\newblock New ideas about temporal pooling.
\newblock Wiki Page, Jan 2014.
\newblock URL \url{http://bit.ly/temporal-pooling}.

\bibitem[Hawkins and Ahmad(2015)]{hawkins2015neurons}
Jeff Hawkins and Subutai Ahmad.
\newblock Why neurons have thousands of synapses, a theory of sequence memory
  in neocortex.
\newblock \emph{arXiv preprint arXiv:1511.00083}, 2015.

\bibitem[Hochreiter and Schmidhuber(1997)]{hochreiter1997long}
Sepp Hochreiter and J{\"u}rgen Schmidhuber.
\newblock Long short-term memory.
\newblock \emph{Neural computation}, 9\penalty0 (8):\penalty0 1735--1780, 1997.

\bibitem[Hutter(2005)]{Hutter:04uaibook}
Marcus Hutter.
\newblock \emph{Universal Artificial Intelligence: Sequential Decisions based
  on Algorithmic Probability}.
\newblock Springer, Berlin, 2005.
\newblock ISBN 3-540-22139-5.
\newblock \doi{10.1007/b138233}.
\newblock URL \url{http://www.hutter1.net/ai/uaibook.htm}.

\bibitem[Ijspeert et~al.(2002)Ijspeert, Nakanishi, and
  Schaal]{ijspeert2002movement}
Auke~Jan Ijspeert, Jun Nakanishi, and Stefan Schaal.
\newblock Movement imitation with nonlinear dynamical systems in humanoid
  robots.
\newblock In \emph{Robotics and Automation, 2002. Proceedings. ICRA'02. IEEE
  International Conference on}, volume~2, pages 1398--1403. IEEE, 2002.

\bibitem[Kato et~al.(2013)Kato, Soriano, Pereda, Fischer, and
  Mirasso]{kato2013limits}
Hideyuki Kato, Miguel~C Soriano, Ernesto Pereda, Ingo Fischer, and Claudio~R
  Mirasso.
\newblock Limits to detection of generalized synchronization in delay-coupled
  chaotic oscillators.
\newblock \emph{Physical Review E}, 88\penalty0 (6):\penalty0 062924, 2013.

\bibitem[Kello et~al.(2007)Kello, Beltz, Holden, and
  Van~Orden]{kello2007emergent}
Christopher~T Kello, Brandon~C Beltz, John~G Holden, and Guy~C Van~Orden.
\newblock The emergent coordination of cognitive function.
\newblock \emph{Journal of Experimental Psychology: General}, 136\penalty0
  (4):\penalty0 551, 2007.

\bibitem[Laukien(2015)]{222464_2015_35283}
Eric Laukien.
\newblock {NeoRL: Neocortex-like reinforcement learning that runs on the GPU or
  CPU (OpenCL)}, December 2015.
\newblock URL \url{http://dx.doi.org/10.5281/zenodo.35283}.

\bibitem[Loh et~al.(2007)Loh, Rolls, and Deco]{loh2007dynamical}
Marco Loh, Edmund~T Rolls, and Gustavo Deco.
\newblock A dynamical systems hypothesis of schizophrenia.
\newblock \emph{PLOS Computational Biology}, 2007.
\newblock URL \url{http://dx.doi.org/10.1371/journal.pcbi.0030228}.

\bibitem[Lorenz(1963)]{lorenz1963}
Edward~N. Lorenz.
\newblock Deterministic nonperiodic flow.
\newblock \emph{Journal of the Atmospheric Sciences}, 20\penalty0 (2):\penalty0
  130--141, 2015/09/26 1963.
\newblock \doi{10.1175/1520-0469(1963)020<0130:DNF>2.0.CO;2}.
\newblock URL
  \url{http://dx.doi.org/10.1175/1520-0469(1963)020<0130:DNF>2.0.CO;2}.

\bibitem[Maass(1997)]{maass1997networks}
Wolfgang Maass.
\newblock Networks of spiking neurons: the third generation of neural network
  models.
\newblock \emph{Neural networks}, 10\penalty0 (9):\penalty0 1659--1671, 1997.
\newblock URL
  \url{http://amath.kaist.ac.kr/~nipl/am621/lecturenotes/spiking_neurons_2.pdf}.

\bibitem[Maass(2000)]{WTAMaass2000}
Wolfgang Maass.
\newblock On the computational power of winner-take-all.
\newblock \emph{Neural Computation}, 12\penalty0 (11):\penalty0 2519--2535,
  2015/09/19 2000.
\newblock \doi{10.1162/089976600300014827}.
\newblock URL \url{http://dx.doi.org/10.1162/089976600300014827}.

\bibitem[Marcus and Williams(2008)]{Marcus:2008}
B.~Marcus and S.~Williams.
\newblock {S}ymbolic dynamics.
\newblock 3\penalty0 (11):\penalty0 2923, 2008.
\newblock URL \url{http://www.scholarpedia.org/article/Symbolic_dynamics}.
\newblock {revision 143845}.

\bibitem[Masulli et~al.(1999)Masulli, Parenti, and Studer]{masulli1999neural}
Francesco Masulli, Riccardo Parenti, and L{\'e}onard Studer.
\newblock Neural modeling of non-linear processes: relevance of the
  takens--man{\'e} theorem.
\newblock \emph{International Journal of Chaos Theory and Applications},
  4\penalty0 (2-3):\penalty0 59--74, 1999.

\bibitem[Mnih et~al.(2013)Mnih, Kavukcuoglu, Silver, Graves, Antonoglou,
  Wierstra, and Riedmiller]{mnih2013playing}
Volodymyr Mnih, Koray Kavukcuoglu, David Silver, Alex Graves, Ioannis
  Antonoglou, Daan Wierstra, and Martin Riedmiller.
\newblock Playing atari with deep reinforcement learning.
\newblock \emph{arXiv preprint arXiv:1312.5602}, 2013.

\bibitem[Ott(2006)]{Ott:2006}
E.~Ott.
\newblock {C}ontrolling chaos.
\newblock 1\penalty0 (8):\penalty0 1699, 2006.
\newblock URL \url{http://www.scholarpedia.org/article/Controlling_chaos}.
\newblock {revision 91167}.

\bibitem[Pecora and Carroll(1990)]{PhysRevLett.64.821}
Louis~M. Pecora and Thomas~L. Carroll.
\newblock Synchronization in chaotic systems.
\newblock \emph{Phys. Rev. Lett.}, 64:\penalty0 821--824, Feb 1990.
\newblock \doi{10.1103/PhysRevLett.64.821}.
\newblock URL \url{http://link.aps.org/doi/10.1103/PhysRevLett.64.821}.

\bibitem[Pikovsky et~al.(2003)Pikovsky, Rosenblum, and
  Kurths]{pikovsky2003synchronization}
Arkady Pikovsky, Michael Rosenblum, and J{\"u}rgen Kurths.
\newblock \emph{Synchronization: a universal concept in nonlinear sciences},
  volume~12.
\newblock Cambridge university press, 2003.

\bibitem[Rulkov et~al.(1995)Rulkov, Sushchik, Tsimring, and
  Abarbanel]{rulkov1995generalized}
Nikolai~F Rulkov, Mikhail~M Sushchik, Lev~S Tsimring, and Henry~DI Abarbanel.
\newblock Generalized synchronization of chaos in directionally coupled chaotic
  systems.
\newblock \emph{Physical Review E}, 51\penalty0 (2):\penalty0 980, 1995.

\bibitem[Sauer(2006)]{Sauer:2006}
T.~D. Sauer.
\newblock {A}ttractor reconstruction, 2006.
\newblock URL
  \url{http://www.scholarpedia.org/article/Attractor_reconstruction}.
\newblock Revision 91017.

\bibitem[Strogatz(2014)]{strogatz2014nonlinear}
Steven~H Strogatz.
\newblock \emph{{N}onlinear {d}ynamics and {c}haos: with applications to
  physics, biology, chemistry, and engineering}.
\newblock Westview press, 2014.

\bibitem[Sugihara et~al.(2012)Sugihara, May, Ye, Hsieh, Deyle, Fogarty, and
  Munch]{sugihara2012detecting}
George Sugihara, Robert May, Hao Ye, Chih-hao Hsieh, Ethan Deyle, Michael
  Fogarty, and Stephan Munch.
\newblock Detecting causality in complex ecosystems.
\newblock \emph{science}, 338\penalty0 (6106):\penalty0 496--500, 2012.

\bibitem[Taga et~al.(1991)Taga, Yamaguchi, and Shimizu]{taga1991self}
Gentaro Taga, Yoko Yamaguchi, and Hiroshi Shimizu.
\newblock Self-organized control of bipedal locomotion by neural oscillators in
  unpredictable environment.
\newblock \emph{Biological cybernetics}, 65\penalty0 (3):\penalty0 147--159,
  1991.

\bibitem[Takens(1981)]{takens1981detecting}
Floris Takens.
\newblock \emph{Detecting strange attractors in turbulence}.
\newblock Springer, 1981.
\newblock URL \url{http://bit.ly/takens-1981}.

\bibitem[Triplett et~al.(2009)Triplett, Owens, Yamada, Lemke, Cang, Stryker,
  and Feldheim]{triplett2009retinal}
Jason~W Triplett, Melinda~T Owens, Jena Yamada, Greg Lemke, Jianhua Cang,
  Michael~P Stryker, and David~A Feldheim.
\newblock Retinal input instructs alignment of visual topographic maps.
\newblock \emph{Cell}, 139\penalty0 (1):\penalty0 175--185, 2009.

\bibitem[Warren(2006)]{warren2006dynamics}
William~H Warren.
\newblock The dynamics of perception and action.
\newblock \emph{Psychological review}, 113\penalty0 (2):\penalty0 358, 2006.

\bibitem[Webber(2015)]{webber2015}
Francisco Webber.
\newblock {Semantic Folding Theory And its Application in Semantic
  Fingerprinting}, 2015.
\newblock URL \url{http://bit.ly/semantic-folding}.
\newblock Cortical.io White Paper.

\bibitem[Whitney(1936)]{whitney1936differentiable}
Hassler Whitney.
\newblock Differentiable manifolds.
\newblock \emph{Annals of Mathematics}, pages 645--680, 1936.

\bibitem[Wijnants et~al.(2012)Wijnants, Hasselman, Cox, Bosman, and
  Van~Orden]{wijnants2012interaction}
ML~Wijnants, F~Hasselman, RFA Cox, AMT Bosman, and G~Van~Orden.
\newblock An interaction-dominant perspective on reading fluency and dyslexia.
\newblock \emph{Annals of dyslexia}, 62\penalty0 (2):\penalty0 100--119, 2012.

\bibitem[Wikimedia(2006)]{wikiLorenz}
Mrubel~Commons Wikimedia.
\newblock A solution in the lorenz attractor rendered as a metal wire to show
  direction and 3d structure, 2006.
\newblock URL
  \url{https://commons.wikimedia.org/wiki/File:Lorenzstill-rubel.png#/media/File:Lorenzstill-rubel.png}.
\newblock File: {\tt Lorenzstill-rubel.png}.

\bibitem[Yamins et~al.(2013)Yamins, Hong, Cadieu, and DiCarlo]{NIPS2013_4991}
Daniel~L Yamins, Ha~Hong, Charles Cadieu, and James~J DiCarlo.
\newblock Hierarchical modular optimization of convolutional networks achieves
  representations similar to macaque it and human ventral stream.
\newblock In C.J.C. Burges, L.~Bottou, M.~Welling, Z.~Ghahramani, and K.Q.
  Weinberger, editors, \emph{Advances in Neural Information Processing Systems
  26}, pages 3093--3101. Curran Associates, Inc., 2013.
\newblock URL \url{http://bit.ly/dl-ventral}.

\end{thebibliography}
}
\bibliographystyle{plainnat}

\end{document}